\def\year{2020}\relax
\documentclass[letterpaper]{article} 
\usepackage{aaai20}  
\usepackage{times}  
\usepackage{helvet} 
\usepackage{courier}  
\usepackage[hyphens]{url}  
\usepackage{graphicx} 
\usepackage{graphicx}  
\title{Geometry Sharing Network for 3D Point Cloud Classification and Segmentation}
\author{Mingye Xu, \textsuperscript{\rm 1 2 }\thanks{M.Xu and Z.Zhou contributed equally.} \quad
	\Large \textbf{Zhipeng Zhou},\textsuperscript{\rm 1 }\footnotemark[1] \quad
	\Large \textbf{Yu Qiao} \textsuperscript{\rm 1 3}\thanks{Corresponding author.}   \\
	\textsuperscript{\rm 1}ShenZhen Key Lab of Computer Vision and Pattern Recognition,\\ SIAT-SenseTime Joint Lab,Shenzhen Institutes of Advanced Technology, Chinese Academy of Sciences\\ 
	\textsuperscript{\rm 2}University of Chinese Academy of Sciences\\ 
	\textsuperscript{\rm 3}SIAT Branch, Shenzhen Institute of Artificial Intelligence and Robotics for Society\\ 
	\{my.xu, zp.zhou, yu.qiao\}@siat.ac.cn 
}

\begin{document}

\maketitle

\begin{abstract}
In spite of the recent progresses on classifying 3D point cloud with deep CNNs, large geometric transformations like rotation and translation remain challenging problem and harm the final classification performance. To address this challenge, we propose Geometry Sharing Network (GS-Net) which effectively learns point descriptors with holistic context to enhance the robustness to geometric transformations. Compared with previous 3D point CNNs which perform convolution on nearby points, GS-Net can aggregate point features in a more global way. Specially, GS-Net consists of Geometry Similarity Connection (GSC) modules which exploit Eigen-Graph to group distant points with similar and relevant geometric information, and aggregate features from nearest neighbors in both Euclidean space and Eigenvalue space. This design allows GS-Net to efficiently capture both local and holistic geometric features such as symmetry, curvature, convexity and connectivity. Theoretically, we show the nearest neighbors of each point in Eigenvalue space are invariant to rotation and translation. We conduct extensive experiments on public datasets, ModelNet40, ShapeNet Part. Experiments demonstrate that GS-Net achieves the state-of-the-art performances on major datasets, 93.3\% on ModelNet40, and are more robust to geometric transformations. Code is released on \url{https://github.com/MingyeXu/GS-Net}.
\end{abstract}

\section{Introduction}

Analysis and classification of 3D point cloud is an important problem in computer vision and graphics, due to its wide applications in robot manipulation \cite{rusu2008towards}, autonomous driving \cite{qi2018frustum} etc. The challenge of this problem comes from several aspects. Firstly, the point cloud are sparsely sampled from 3D surfaces in an irregular and off-order way. Secondly, the point cloud usually undergoes large geometric transformations and deformations. It is important to achieve robustness to transformation and permutation for analyzing and classifying 3D point cloud.

\begin{figure}[t]
	\centering
	\includegraphics[width=1\columnwidth]{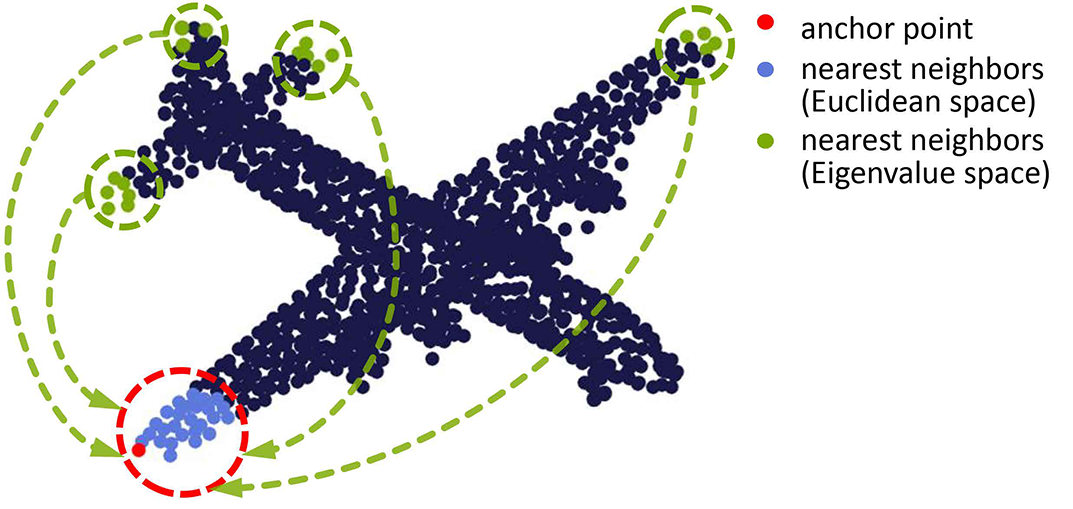} 
	\caption{Visualization of Eigen-Graph (Best view in color and zoom in). Given a red anchor point, traditional convolutions for 3D points operate on a local region as shown in the red circle. To explore the geometry of point cloud such as symmetry, curvature, convexity and connectivity, we use Eigen-Graph to group neighbors in both Euclidean space and Eigenvalue space. The anchor point's neighbors in Euclidean space are colored blue and its neighbors in Eigenvalue space are colored green, thus we construct the Eigen-Graph of the anchor point and its neighbors. Obviously, green points provide more information about the geometry of the whole point cloud. It shows that our method indeed associates the anchor points with points having similar local geometry, even though these points are far away from each other in Euclidean space.}
	\label{fig_eigenGraph}
\end{figure}

Large  research efforts have been devoted to solving the above problems.
One research direction \cite{su2015multi,feng2018gvcnn,Maturana2015VoxNet,Wu20153D} aims to represent the irregular 3D point cloud using regular data, in that way they can use classical convolution neural network to process the regular data. Two of the most common regular representations are voxels and multi-view images. However, both these representations have limitations. Dense voxels representation is inefficient due to the sparsity of input point clouds, while multi-view images may lose 3D structures of points and cause occlusion problem.

Another direction focuses on designing convolution operations for irregular points, which are inspired by the prominent success of CNNs on regular grid data, such as audio and images. 
PointNet \cite{qi2017pointnet} learns a spatial encoding of each point directly on Euclidean space and aggregates all individual point features by max pooling to obtain a global point cloud signature. Max pooling as a symmetric operation can obtain permutation invariance. By its design, PointNet does not capture local geometry directly which is indispensable to the description of 3D shape.
Other works \cite{qi2017pointnet++,xu2018spidercnn} mainly utilize group operations (k-nearest neighbors group or ball region group) to identify local points for convolution.
But these group operations only focus on local neighborhood region in Euclidean space.
Despite of the discreteness and irregularity of point cloud, these operations mainly account the local structures of each point and are not efficient to capture the holistic geometric information from distant information. The holistic geometry not only provides discriminative cues for classification but also help to achieve robustness to transformation. In addition, points with similar geometric structures can be far away from each other in Euclidean space. Previous works mentioned above largely neglect the geometric relationships among these distant points.

Inspired by the above analysis, this paper proposes Geometry Sharing Network (GS-Net) which aggregates features in both Euclidean space and Eigenvalue space. GS-Net exploits Eigen-Graph to calculate structure tensor for measuring local geometric properties of input points, which further allows us to identify points with similar local structures but located distant in Euclidean space. We prove that these structure tensors are invariant to transformations and yield rich local structural information. As shown in Figure \ref{fig_eigenGraph}, given an anchor point for convolution, GS-Net identifies a group of neighbor points from Euclidean space and also another group of points with similar local structures (Eigen-graph features). Then the convolutions are performed for both groups to capture local and holistic geometric representation separately. The convolutional features from both groups are integrated for classification or segmentation. We conduct extensive experiments to examine the proposed methods. Our method achieves the state-of-the-art performance on ModelNet40 for classification (93.3\%) and shows more robustness to geometric transformations than previous methods.

The main contributions of this paper are summarized as follows. 
\begin{itemize}
	\item  We propose a novel Geometry Similarity Connection (GSC) module which exploits Eigen-Graph to group distant points with similar and relevant geometric information and aggregate features from neighbors in both Euclidean space and Eigenvalue space which can capture local and holistic geometric information more efficiently.
	\item We introduce 3D structure tensor and Eigen-Graph to capture the geometric features of points. Theoretically, we prove these features are invariant to translation and rotation. 
	\item Our GS-Net achieves the state-of-the-art performances on major datasets, ModelNet40, ShapeNet Part. Moreover, GSC module can be integrated into different existing pipelines for point cloud analysis.
\end{itemize}

\section{ Related Work}
\subsection{ Deep Learning on Point Cloud Analysis}

Deep neural networks have enjoyed remarkable success for various vision tasks, however it remains challenging to apply CNNs to domains lacking a regular structure such as 3D point cloud. 
These challenges include: (1) local and holistic geometric information representation; (2) permutation invariance; (3) rotation and translation invariance.
However, not all networks can address these problems absolutely.

PointNet \cite{qi2017pointnet} and DeepSet \cite{Zaheer2017Deep} are pioneering architectures that directly process point cloud. The basic idea is to learn a spatial encoding of each point and then aggregate all individual point features to a holistic signature. But by this design, relations between points are not sufficiently captured. 
To remedy this, PointNet++ \cite{qi2017pointnet++} partitions point cloud into overlapping local regions by the distance metric of the underlying space and extracts local features capturing fine geometric structures from neighbors, but it still only considers every point in its local region independently. In our method, we address this issue by defining a convolution block that group the features from the neighbors in Euclidean space and Eigenvalue space. 

DGCNN \cite{wang2018dynamic} captures local geometric structure while maintaining permutation invariance and reconstructs the $k$-nn graph using nearest neighbors in the features space produced by each layer. Different with DGCNN, our method does not use dynamic strategy, we apply Eigen-Decomposition to choose the nearest neighbors and share the local features with distant points with similar geometric information. 

\cite{thomas2018semantic} directly uses eigenvalues and fuctions of eigenvalues as features in deep-learning setting. \cite{landrieu2018large} adds eigenvalues to its shape descriptors. Our method aggregates features from nearest neighbors in Eigenvalue space in order to capture holistic geometric information. And we also uses eigenvalue as features in our network settings.  

\subsection{Classical Geometric Representation }
The local geometry of point cloud is estimated by the distribution of points in the neighborhood. \cite{demantke2011dimensionality} propose a method which aims at finding the optimal neighborhood radius for each point, working directly and exclusively in the 3D domain, without relying on surface descriptors or structures. Firstly, they compute three dimensionality features for each point, between predefined minimal and maximal neighborhood scale. The three dimensionality features $(a_{1D},a_{2D},a_{3D})$ \cite{demantke2011dimensionality} are computed exhaustively, at each point and for each accepted neighborhood scale from local covariance matrix. Various geometrical features can be derived from the eigenvalues of the covariance matrix. $a_{1D},a_{2D},a_{3D}$ describe linear, planar, and scatter respectively. 
In our GS-Net, we use operations on eigenvalues to improve the robustness of rotation and translation in GS-Net. Moreover, the Eigen-Graph (Sec \ref{sec_GSCM}) enhances the representation of local geometry. 

\subsection{Rotation Invariance for Point Cloud Analysis }
In comparison to permutation invariance, rotation invariance is a more challenging problem. Previous works has dealt with issues of invariance or equivarance under particular input transformations. PointNet \cite{qi2017pointnet} and PointNet++ \cite{qi2017pointnet++} guarantee the permutation invariance by a symmetric pooling operator and PointNet employs a complex and computationally intensive spatial transformer network to learn 3D alignment, PCPNet \cite{guerrero2018pcpnet} also uses a learned transformer block, but these networks (including \cite{wang2018dynamic,xu2018spidercnn}) do not include rotation invariance. \cite{Thomas2018Tensor} upgrades the existing neural network with rotation invariance property, a special convolutional operation is designed as a basis block in the network. But it causes the loss of information as there is no bijection between $R^3$ and 2-dimensional sphere. In our method, Eigen-Graph addresses the rotation invariance naturally with Eigen-Decomposition of 3D structure tensor.

\section{Method}
\subsection{Overview}

\begin{figure*}[t]
	\centering
	\includegraphics[width=2.1\columnwidth]{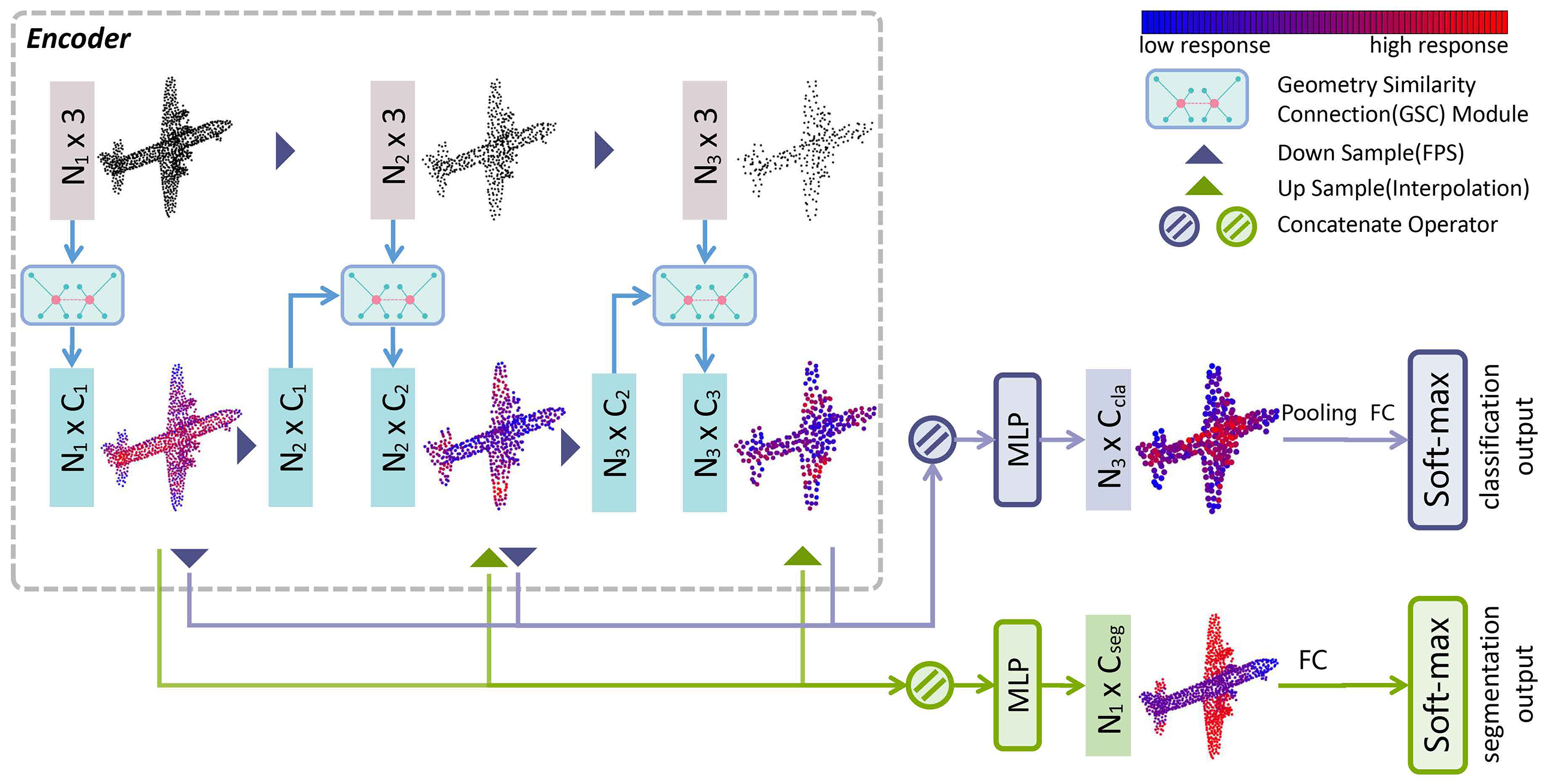} 
	\caption{GS-Net architecture for classification and segmentation (Best view in color and zoom in). In encoder, we use hierarchical structure to learn the features on each level. The network takes N points as input. We use Geometry Similarity Connection (GSC) module to capture abundant local geometry features of each point which can be shared with distant points. After that, we down-sample the points and features to the next level. For classification, we concatenate down-sampled features from each level and pool the holistic features to a 1D global descriptor, which is used to generate classification scores. The segmentation network concatenate the interpolated features of each levels and then calculate each point's scores. The color from red to blue denotes the features' response, while the size of points denotes the features' channels number on different levels.}
	\label{fig1}
\end{figure*}

As shown in Figure \ref{fig1}, we consider a $S$-dimensional point cloud with $N$ points, denoted by $\mathcal{X}=\left\{x_{1}, \ldots, x_{N}\right\} \subset \mathbf{R}^{S}$. Usually, each point of point cloud contains 3D coordinates $x_{i}=\left(x_{i}^{1}, x_{i}^{2}, x_{i}^{3}\right)$, which means that $S = 3$; it is also possible to include other coordinates representing RGB information, normal vectors, and so on.
In our network architecture, we use hierarchical structure to learn local and holistic features of point cloud.  On each level, we use Geometry Similarity Connection(GSC) module (Sec \ref{sec_GSCM}) to capture abundant local geometric information of each point and share geometric features with distant points. After that, we adopt the FPS algorithm to down-sample the points and the features (Sec \ref{orther_design}). Low-level features represent the local geometric information, while high-level features provide semantic information.

As for classification task, instead of using only the last level's features as the encoder's output \cite{wang2018dynamic}, we concatenate all levels' features together and extract the holistic features by global max pooling and global average pooling. The concatenation of all levels' features aims to fuse the features from different levels and the pooling operator urges to capture the most effective features for classification. Then we handle the holistic features by fully-connected layers with integrated dropout \cite{Srivastava2014Dropout} to calculate the probability for each category. The cross-entropy loss is used for training.

As for segmentation task, our segmentation network has an encoder which is the same as the classification network's. We need to interpolate the features on each level of the encoder module and then concatenate them. Inspired by \cite{qi2017pointnet++}, we also concatenate repeated one-hot category label to the features before MLP \cite{hornik1991approximation}. This mechanism is designed to apply the category supervision to the point-wise segmentation.

\subsection{ Geometry Similarity Connection Module
\label{sec_GSCM}}
\begin{figure}[t]
	\centering
	\includegraphics[width=1\columnwidth]{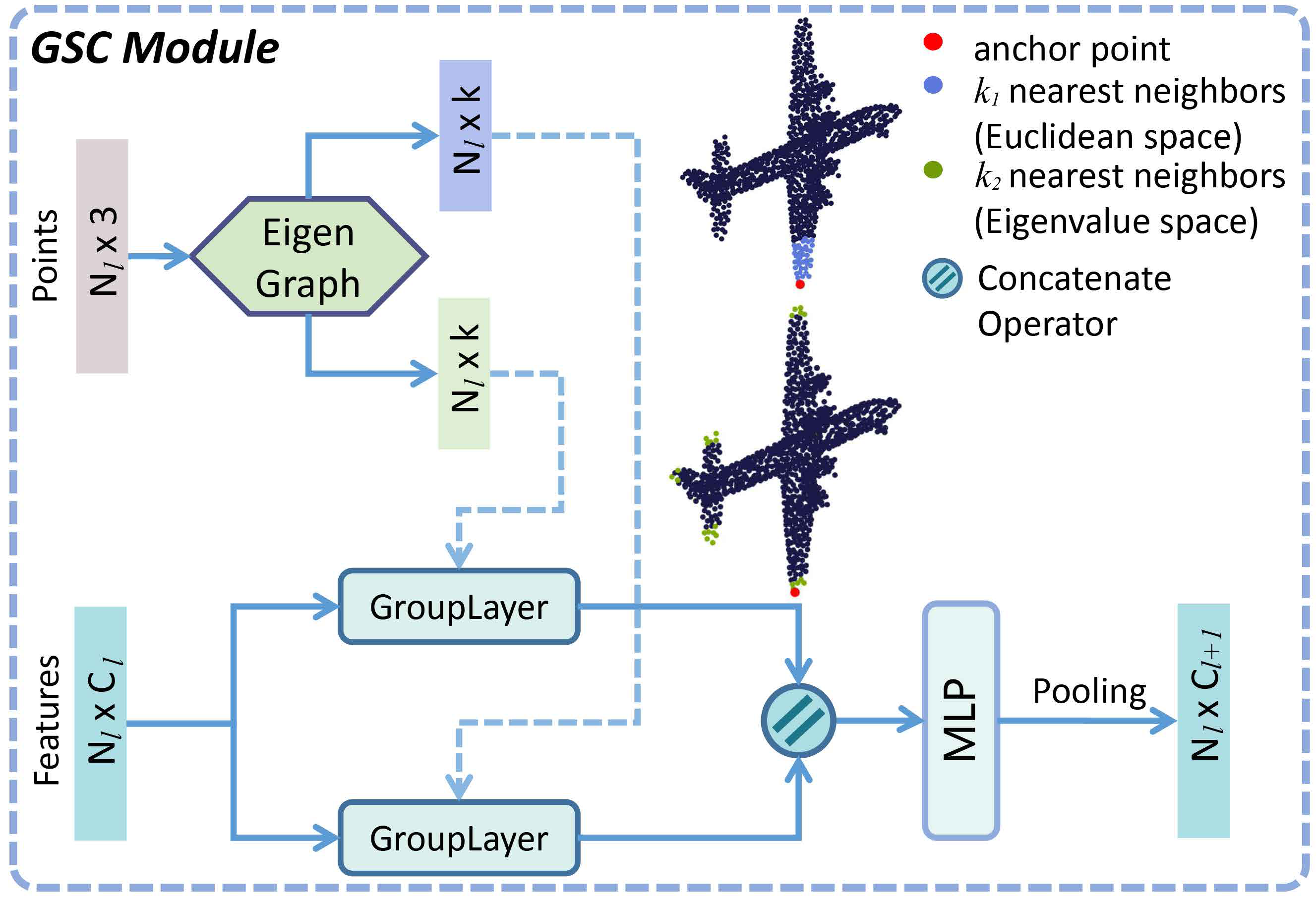} 
	\caption{Geometry Similarity Connection module (Best view in color and zoom in). $k_1$ denotes the number of nearest neighbors in Euclidean space and $k_2$ means number of nearest neighbors in Eigenvalue space. The input are points and features on level-$l$. In GSCM, Eigen-Graph is designed to compute the indices of neighbors in Euclidean space and Eigenvalue space. The GroupLayer shares the local geometric features according to the indices of $k_2$-nearest neighbors in Eigenvalue space.    }
	\label{fig2}
\end{figure}

\begin{figure}[t]
	\centering
	\includegraphics[width=1\columnwidth]{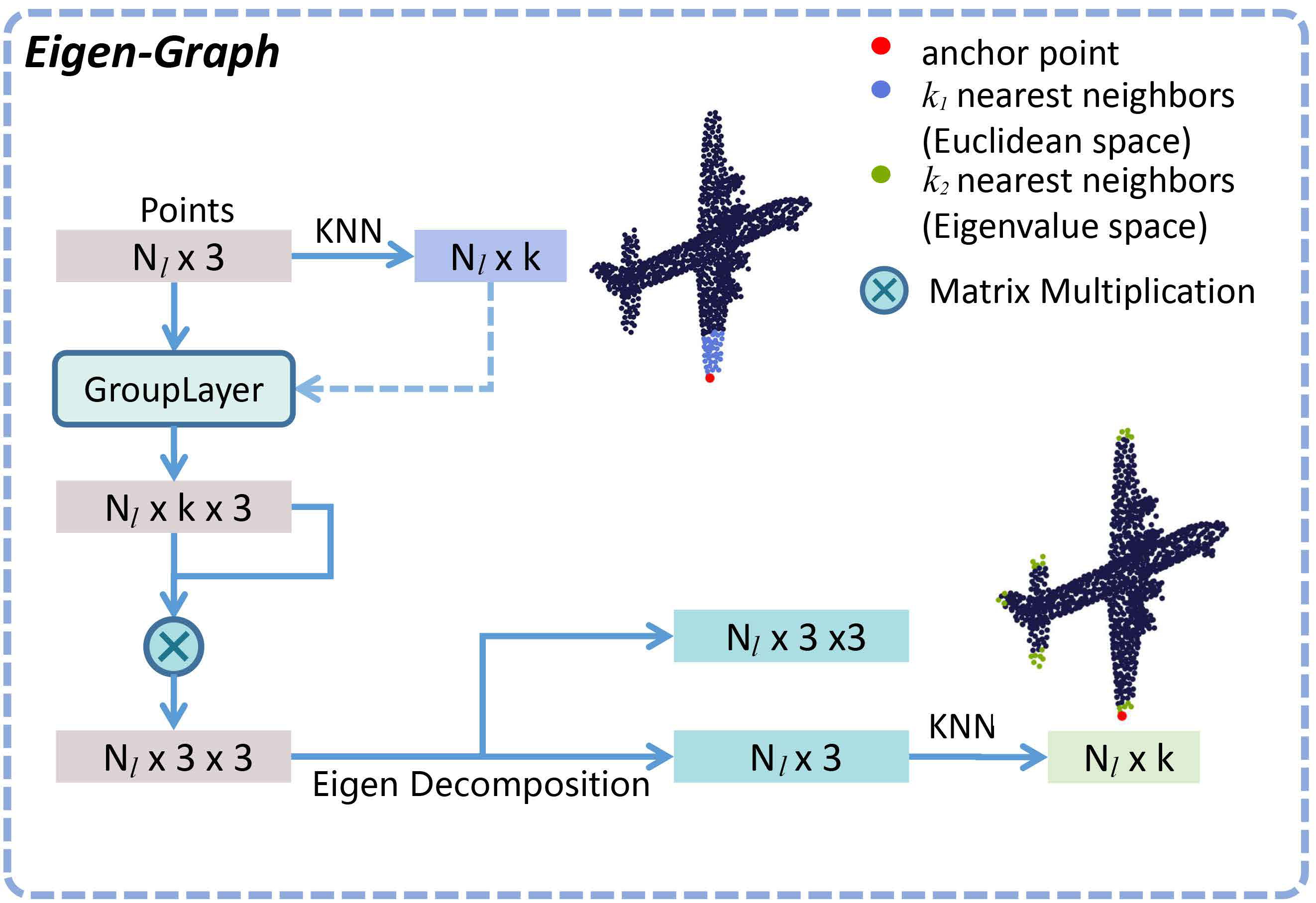} 
	\caption{Eigen-Graph module(Best view in color and zoom in). It first gets the structure tensor of the input points, then use Eigen-Decomposition to compute the eigenvalues which are used to get the indices of the nearest neighbors in Eigenvalue space.}
	\label{fig3}
\end{figure}

This subsection describes the Geometry Similarity Connection (GSC) module in GS-Net. It is illustrated in Figure \ref{fig2} and \ref{fig3}. The structure of Eigen-Graph is shown in Figure \ref{fig_eigenGraph}.

\textbf{Eigen-Graph.}
As shown in Figure \ref{fig3}, we use $k$-nearest neighbors search (KNN) algorithm to get $k_1$-nearest neighbors of each point $x_i$ in Euclidean space. Let $\left\{x_{i_{1}}, \ldots, x_{i_{k_{1}}}\right\}$ be $k_1$-nearest neighbors of $x_i$. Let $M=\left(x_{i_{1}}-x_{i}, \ldots, x_{i_{k_{1}}}-x_{i}\right)$, where $x_{i_{j}}(1\leq j \leq k_1)$ belongs to $k_1$-nearest neighbors of $x_i$ in Euclidean space.

We define the 3D structure tensor as $C=MM^T$, even if the ground truth (surface) is locally flat, noise points cause unflatness of point cloud sampled from the surface. As long as the neighbor region of the given point is not flat, $C$ is a symmetric positive definite matrix. We have the decomposition $C=R \Lambda R^{T}$, where $R$ is a rotation matrix and $\Lambda$ is a diagonal and positive definite matrix, known as eigenvectors and eigenvalues matrices respectively. 
The positive eigenvalues $\lambda \subset \mathbf{R}^{3}$ are ordered so that $\lambda^{1} \geq \lambda^{2} \geq \lambda^{3}>0$. 
At each point $x_i$, we get the 3D structure tensor and denote the eigenvalues at point $x_i$ by $\left(\lambda_{i}^{1}, \lambda_{i}^{2}, \lambda_{i}^{3}\right) (1 \leq i \leq N)$. We use $L^{2}$ norm to calculate the distances between different points. 

\begin{equation}
{Distance}(x_{i}, x_{j})=\|\lambda_{i}-\lambda_{j}\|_{L^{2}}
\end{equation}

We choose the indices of $k_2$-nearest neighbors of each point according to Eigen Matrix whose element is  $D_{i j}=\left\|\lambda_{i}-\lambda_{j}\right\|_{L^{2}}$.

\textbf{GroupLayer.}
Now we have $k_1$-nearest neighbors' indices in Euclidean space and $k_2$-nearest neighbors' indices in Eigenvalue space. As we have presented in Figure \ref{fig2}, we denote the input features of level $l$ by $F^l = \{f_1^l,f_2^l,...,f_N^l\}$. For convenience, we omit the superscript $l$. In GroupLayer, let  $\left\{f_{i_{1}^{1}}, \dots, f_{i_{k_{1}}^{1}}\right\}$ be $k_1$-nearest neighbors' features of point $x_i$, and let $\left\{f_{i_{1}^{2}}, \dots, f_{i_{k_{2}}^{2}}\right\}$ be $k_2$-nearest neighbors' features of point $x_i$. We group the neighbor features as follows:
\begin{equation}
f_{i}^{k_1}=\bigodot_{j :(i, j) \in \mathcal{E}}\left(f_{j}-f_{i}, f_{j}\right), j \in\left\{i_{1}^{1}, \ldots, i_{k_{1}}^{1}\right\} \end{equation}

\begin{equation}
f_{i}^{k_2}=\bigodot_{p :(i, p) \in \mathcal{E}}\left(f_{p}-f_{i}, f_{p}\right), p \in\left\{i_{1}^{2}, \ldots, i_{k_{2}}^{2}\right\}
\end{equation}
where $\bigodot$ means concatenation.
Then we concatenate $f_{i}^{k_1}$ with $f_{i}^{k_2}$ as the features at each point:
\begin{equation}
f_{i}^{\prime}=\bigodot\left(f_{i}^{k_1},f_{i}^{k_2}\right)
\end{equation}

In the first GSC module shown in Figure \ref{fig1}, the input features are the coordinates $X$ and the eigenvalues $E$ of points. We group coordinates using $k_1$-nearest neighbors and group eigenvalues using $k_2$-nearest neighbors. In the other GSC modules, we use the previous level's output as the input features and group features in both Euclidean space and Eigenvalue space.

\begin{table*}
	\scriptsize
	\begin{center}
		\begin{tabular}{p{123pt} | p{21pt} | p{9pt} p{9pt} p{9pt} p{9pt} p{9pt} p{9pt} p{9pt} p{9pt} p{9pt} p{9pt} p{9pt} p{9pt} p{9pt} p{9pt} p{9pt} p{9pt} }
			\hline
			method& instance & aero & bag & cap & car & chair & ear & guitar & {knife} & lamp & lap & motor & mug & pistol & rocket & { skate} & table  \\
			& m-IOU  &   &   &   &   &   & ph &   &   &   & top   &   &   &  &   & board & \\  
			\hline
			Kd-Net\cite{klokov2017escape} & 82.3 & 80.1 & 74.6 & 74.3 & 70.3 & 88.6 & 73.5 & 90.2 & 87.2 & 81.0 & 94.9 & 57.4 & 86.7 & 78.1 & 51.8 & 69.9 & 80.3\\
			PointNet\cite{qi2017pointnet} & 83.7 & 83.4 & 78.7 & 82.5 & 74.9 & 89.6 & 73.0 & {91.5} & 85.9 & 80.8 & 95.3 &  65.2 & 93.0 & 81.2 & 57.9 & 72.8 & 80.6 \\
			SCN\cite{xie2018attentional} &84.6  & 83.8 & 80.8 & 83.5 & \textbf{79.3} & 90.5 & 69.8 & \textbf{91.7} & 86.5 & 82.9 & \textbf{96.0} & {69.2} & 93.8 & 82.5 & \textbf{62.9} & 74.4 & 80.8\\
			SO-Net\cite{li2018so} & 84.6 & 81.9 & 83.5 & 84.8 & 78.1 & 90.8 & 72.2 & 90.1 & 83.6 & 82.3 & 95.2 & 69.3 & 94.2 & 80.0 & 51.6 & 72.1 & 82.6\\
			KCNet\cite{shen2018mining} &84.7  & 82.8 & 81.5 & 86.4 & 77.6 & 90.3 & 76.8 & 91.0 & 87.0 & \textbf{84.5} & 95.5 & 69.2 & 94.4 & 81.6 & 60.1 & 75.2 & 81.3\\
			RS-Net\cite{huang2018recurrent} & 84.9 & 82.7 & \textbf{86.4} & 84.1 & 78.2 & 90.4 & 69.3 & 91.4 & 87.0 & 83.5 & 95.4 & 66.0 & 92.6 & 81.8 & 56.1 & 75.8 & 82.2\\
			PointNet++\cite{qi2017pointnet++} & 85.1 & 82.4 & 79.0 & 87.7 & 77.3 & 90.8 & 71.8 & 91.0 & 85.9 & {83.7} & 95.3 & {71.6} & 94.1 & 81.3 & 58.7 & 76.4 & 82.6  \\
			DGCNN\cite{wang2018dynamic} & 85.1 & \textbf{84.2} & 83.7 & 84.4 & 77.1 & \textbf{90.9} & \textbf{78.5} & {91.5} & \textbf{87.3} & 82.9 & \textbf{96.0} & 67.8 & 93.3 & 82.6 & 59.7 & 75.5 & 82.0 \\
			SpiderCNN\cite{xu2018spidercnn} & 85.3 & 83.5 & 81.0 & 87.2 & 77.5 & 90.7 & 76.8 & 91.1 & \textbf{87.3} & 83.3 & 95.8 & 70.2 & 93.5 & 82.7 & 59.7 & 75.8 & 82.8 \\
			\textbf{Ours} & \textbf{85.3} & 82.9 & {84.3} & \textbf{88.6} & {78.4} & 89.7 & {78.3} & \textbf{91.7} & 86.7 & 81.2 & 95.6 & \textbf{72.8} & \textbf{94.7} & \textbf{83.1} & {62.3} & \textbf{81.5} & \textbf{83.8}\\
			\hline
		\end{tabular}
	\end{center}
	\caption{Segmentation results on ShapeNet Part dataset.}
	\label{Result_sgm}
\end{table*}

\textbf{MLP and MaxPooing.}
In GSC module, we calculate features at each point from GroupLayer and implement the multilayer perception (MLP)\cite{hornik1991approximation}, then we use Max-Pool in neighbor domains to get the features of each point:

\begin{equation}
f_{i}^{\prime \prime }={MaxPool}(M L P\left(f_{i}^{\prime }\right) )
\end{equation}
And the output of GSC module is denoted by $F^{\prime \prime}= \{f_1^{\prime \prime},f_2^{\prime \prime},...,f_N^{\prime \prime}\}$.


\subsection{Rotation and Translation Invariance
	\label{sec_rotation_invariance}
}
In this subsection we give some theoretical analysis about rotation and translation invariant robustness of our method. As we have mentioned in Sec \ref{sec_GSCM}, the 3D structure tensor is $C=MM^T$. We denote the eigenvalues of $C$ as $(\lambda^1,\lambda^2,\lambda^3)$ and the corresponding eigenvectors are $(v^1,v^2,v^3)$. Thus we have the following equation:
\begin{equation}
Cv^q = \lambda^qv^q\ ,  1\leq q \leq 3
\end{equation}

The way we get 3D structure tensor guarantees that 3D structure tensor of each point is invariant to translation. Let $R$ be an arbitrary rotation matrix in 3D Euclidean space. After applying rotation matrix to point cloud, we get the new 3D structure tensor $C' = RM(RM)^T$. We can get the following equations:
\begin{equation}
R^TR = RR^T = I
\end{equation}
\begin{equation}
C'Rv^q = RMM^TR^TRv^q = RMM^Tv^q = RCv^q = \lambda^qRv^q
\end{equation}
From equations above, we know that $(\lambda^1, \lambda^2, \lambda^3)$ are also the eigenvalues of 3D structure tensor $C'$.
So the eigenvalues of each point is invariant to rotation and translation which ensure the indices of $k_2$-nearest neighbors of each point are invariant (illustrated in Figure \ref{rotate_k2}). This mechanism improves the robustness of our model to rotation and translation. 
The empirical experiment results also demonstrate what we have proved theoretically (Sec \ref{robustness_exp}).

\subsection{Complements of the Architecture  
	\label{orther_design}
}
\textbf{Hierarchical Feature Learning.
}
Our method follows the design where the hierarchical structure \cite{qi2017pointnet++} is composed of a set of abstract layers. By this way, we can enlarge receptive field of each point progressively along the hierarchy.
As shown in Figure \ref{fig1}, the hierarchical structure is composed of three abstract levels. An abstract level $l$ takes $N_l \times 3$ points matrix and $N_l \times C_l$ features matrix as input. The output are $N_{l+1} \times 3$ points matrix and $N_{l+1} \times C_{l+1}$ features matrix. We use FPS algorithm to down-sample the points and features at 3 levels (1024-512-256 points in classification network).

\textbf{Feature Interpolation for Segmentation Task.}
In segmentation task, to obtain the feature map which has the same number of points as the original input, we must interpolate features from the coarsest scale to the original scale \cite{qi2017pointnet++}.
The $l$-th features interpolation level takes $N_{l} \times C_{l}^{’}$ decoder features matrix as input, let $X_{l}$ and $X_{1}$ be the spatial points set with $N_{l} \times 3$ and $N_{1} \times 3$ coordinates. To obtain the features of $ 1 $-st level, we simply find three nearest neighbors of $X_{1}$ in $X_l$ and then calculate the weighted sum of their features. The combination weights are acquired according to the neighbors' normalized spatial distances.

\section{Experiments}
In this section, we conduct comprehensive experiments to evaluate our GS-Net. In Sec \ref{Sec_PTAnalysis}, we evaluate our GS-Net for point cloud analysis on classification task and segmentation task. In Sec \ref{robustness_exp}, we compare the rotation robustness of GS-Net with state-of-the-art methods. 
\subsection{Point Cloud Analysis
\label{Sec_PTAnalysis}}

\begin{table}
	\begin{center}
		\scriptsize
		\begin{tabular}{p{138pt} p{39pt} p{20pt}}
			\hline
			Method& Input& Accuracy\\
			\hline
			Pointwise-CNN\cite{Hua2017Point} & 1k points & 86.1\\
			ECC\cite{Simonovsky2017Dynamic} & 1k points & 87.1\\
			PointNet\cite{qi2017pointnet} & 1k points & 89.2\\
			SCN\cite{xie2018attentional} & 1k points  & 90.0 \\
			Flex-Conv\cite{Groh2018Flex} & 1k points  & 90.2\\
			PointNet++\cite{qi2017pointnet++} & 1k points  &  90.7\\ 
			SO-Net\cite{li2018so} & 2k points & 90.9 \\
			KCNet\cite{shen2018mining} & 1k points  & 91.0  \\ 
			MRTNet\cite{Gadelha2018Multiresolution} & 1k points  & 91.2 \\ 
			Spec-GCN\cite{Chu2018Local} & 1k points  & 91.5 \\
			PAT(FPS+GSS)\cite{yang2019modeling} & 1k points  & 91.7 \\
			Kd-Net\cite{klokov2017escape} & 1k points  & 91.8 \\
			SpiderCNN\cite{xu2018spidercnn} & 1k points & 92.2 \\
			DGCNN\cite{wang2018dynamic} & 1k points & 92.2 \\
			PCNN\cite{Atzmon2018Point} & 1k points  & 92.3 \\ 
			\textbf{Ours}  & \textbf{1k points} & \textbf{92.9}\\
			\textbf{Ours(k=32)}  & \textbf{2k points} & \textbf{93.3}\\
			\hline
			PointNet++\cite{qi2017pointnet++} & 5k points+nor& 91.9\\
			Spec-GCN\cite{Chu2018Local} & 1k points+nor & 91.8\\
			SpiderCNN\cite{xu2018spidercnn} & 1k points+nor & 92.4\\
			\hline
		\end{tabular}
	\end{center}
	\caption{Classification results (\%) on ModelNet40 dataset. "nor" denotes the normal of point cloud.}
	\label{Result_classification}
	
\end{table}

\textbf{Classification on ModelNet40.}
ModelNet40 \cite{Wu20153D} contains 12,311 CAD models from 40 categories. 9,843 models are used for training and 2,468 models are for testing. We evaluate our model on the ModelNet40 \cite{Wu20153D} for classification task. Following the configuration in PointNet \cite{qi2017pointnet}, we use the source code of PointNet to sample points uniformly from the mesh models.
The results are summarized in Table \ref{Result_classification}. Our model achieves the state-of-the-art performance (93.3\%).


\textbf{Part Segmentation on ShapeNet Part.}
Part segmentation task is a challenging task for fine-gained shape analysis. We evaluate our method for this task on ShapeNet Part benchmark \cite{yi2016scalable}. ShapeNet Part consists of 16,880 models from 16 shape categories and 50 different parts in total, with 14,006 models for training and 2,874 models for testing split. Each point cloud is annotated with 2 to 6 parts. We choose mIoU as the evaluation metric which is averaged across all classes and instances.
The results are summarized in Table \ref{Result_sgm}. The input consists of coordinates and normals.
Our method can effectively deal with point clouds with geometric characteristic such as symmetrical structure. Figure \ref{fig_seg_vis} shows some segmentation examples.

\subsection{Comparison of Rotation Robustness
\label{robustness_exp}}
We compare GS-Net with the state-of-the-art approaches on ModelNet40 classification for rotation-robustness evaluation. The results are summarized in Table \ref{Result_rotation} with four comparisons: (1) both training set and test set are augmented by random angle rotation for z axis(z/z); (2) training set with random angle rotation for z axis and test set with random angles rotation for all three axes (x,y,z) (z/s); (3) both training set and test set are augmented by random angles rotation for all three axes (s/s); (4) only test set with random angles rotation for all three axes (0/s). 

Table \ref{Result_rotation} consists of two groups of approaches. The first group consists of four approaches: DGCNN \cite{wang2018dynamic}, Point \cite{qi2017pointnet}, Point++ \cite{qi2017pointnet++} and SpiderCNN \cite{xu2018spidercnn}, while the second group is our approach with different settings. Different from our model shown in Figure \ref{fig1}, last one of second group only use eigenvalues as the input features without any coordinates information and it achieves the best performances of comparison (2) and (4). While our original model achieves the best performance of comparison (3) and get a comparable result with DGCNN of comparison (1). These comparisons aim to validate the eigenvalues of each point is invariant to rotation and can improve robustness of our method to rotation.

\begin{figure}[t]
	\centering
	\includegraphics[width=1\columnwidth]{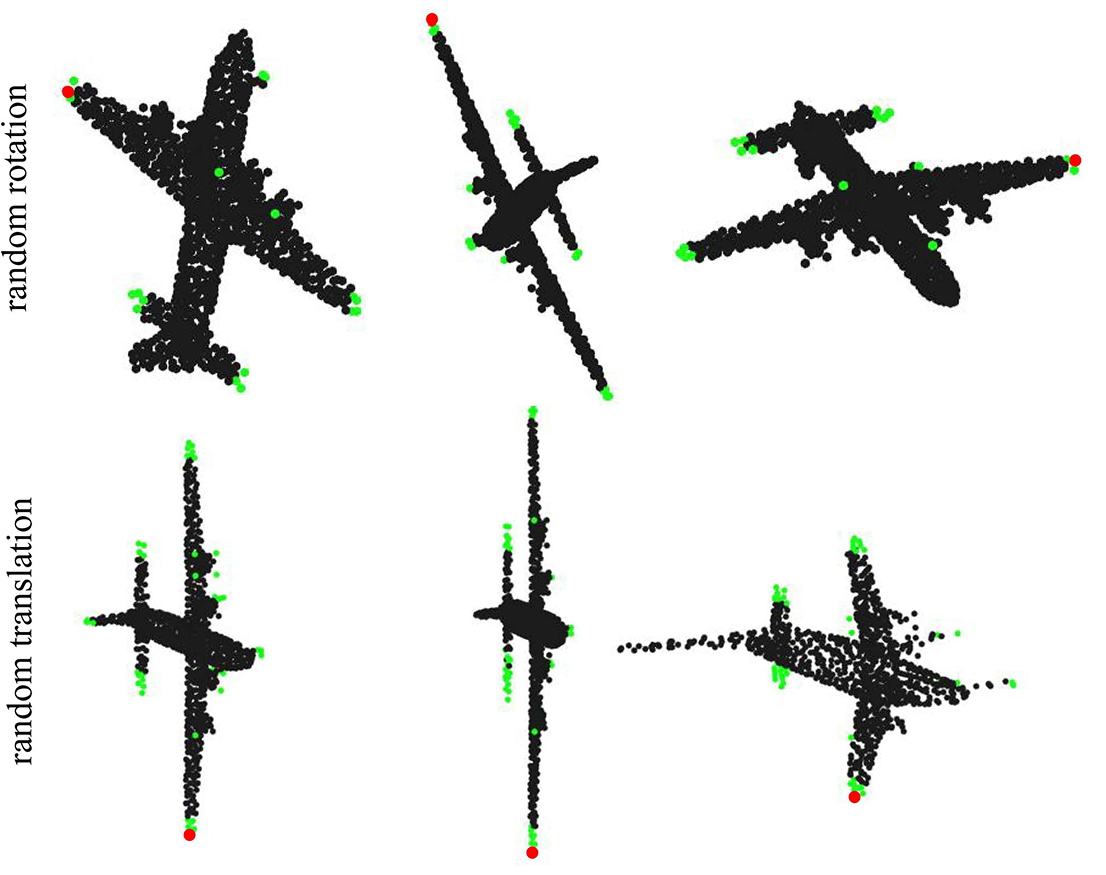} 
	\caption{Visualization of rotation and translation invariance. We visualize the anchor point(red)'s neighbors (green) in Eigenvalue space. The green points are not influenced by the rotation and translation of the input points.  }
	\label{rotate_k2}
\end{figure}

\begin{table}
	\centering
	\footnotesize
	\begin{tabular}{p{110pt}llll}
		\hline
		Method  & z/z & z/s & s/s & 0/s\\
		\hline
		DGCNN\cite{wang2018dynamic} &  \textbf{90.4} & 30.9 & 82.6 & 19.6 \\
		PointNet\cite{qi2017pointnet} &  81.6 & 15.8 & 66.3 & 11.7 \\
		PointNet++\cite{qi2017pointnet++} &  90.1 & 27.1 & 87.8 & 16.1 \\
		SpiderCNN\cite{xu2018spidercnn} &  83.5 & 29.3  & 69.6 & 21.8 \\
		\hline
		\textbf{Ours}        & {89.8} & {37.1} & \textbf{87.9} & {19.4} \\
		\textbf{Ours($\lambda_j - \lambda_i,\lambda_j$)} & {85.0} & \textbf{72.8} & {82.8} & \textbf{50.7} \\
		\hline
	\end{tabular}
	\caption{Comparisons of rotation robustness on ModelNet40 classification. $\lambda_i $ means the eigenvalues of the anchor point and $\lambda_j$ denotes the eigenvalues of its neighbors.}
	\label{Result_rotation}
	
\end{table}

\begin{figure}[t]
	\centering
	\includegraphics[width=1\columnwidth]{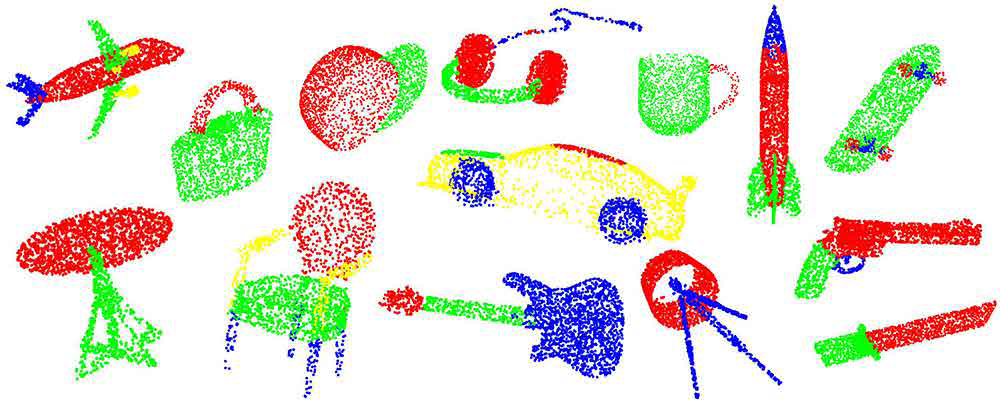} 
	\caption{Segmentation examples on ShapeNet Part.}
	\label{fig_seg_vis}
\end{figure}

\section{Analysis of GS-Net}
In Sec \ref{Sec_abation_ana}, we perform the ablation analysis of GS-Net. We discuss the effectiveness of architecture design and input features. Sec \ref{Sec_ComplexityAnalysis} is the complexity comparison of GS-Net and existing methods. Sec \ref{vis_GSNet} shows that Eigen-Graph efficiently capture local and holistic geometric features such as symmetry and connectivity.
\subsection{Ablation Analysis
\label{Sec_abation_ana}}

\textbf{Analysis of Architecture Design.} 
We analyze the effectiveness of our method's components on ModelNet40 benchmark for classification task. The results are summarized in Table \ref{ablation_study}. All experiments in the ablation study are conducted using $k=20$ nearest neighbors. 

\begin{table}
	\centering
	\begin{tabular}{llll}
		\hline
		FPS & $k$-nn space & \# Points & Accuracy(\%)\\
		\hline
		On & EU + EI & 1024 & 92.8 \\
		Off& EU + EI & 1024 & 92.5 \\
		On & EU + EI & 2048 & 92.9 \\
		On & EU             & 1024 & 92.6 \\
		On & EI            & 1024 & 92.5 \\
		\hline
	\end{tabular}
	\caption{ablation study of architecture design (\%). 'FPS' indicates whether to use FPS down-sample strategy in the classification network. 'EU' denotes that our method groups the neighbors in Euclidean space, while 'EI' denotes that our method groups the neighbors in Eigenvalue space. }
	\label{ablation_study}
\end{table}

\begin{table}
	\centering
	\begin{tabular}{lll}
		\hline
		Input features & Channels & Acc.\\
		\hline
		$x_j$                          & 3    & 92.5 \\
		$\lambda_j$                    & 3    & 85.2 \\
		$s_i$                          & 8    & 91.9 \\
		$x_j-x_i$                      & 3    & 92.6 \\
		$x_j-x_i,x_j$                  & 6    & 92.7 \\
		$x_j-x_i,x_j,\lambda_j-\lambda_i,\lambda_j$      & 12   & 92.8 \\
		$x_j-x_i,x_j,\lambda_j-\lambda_i,\lambda_j,d_{ij}$ & 13   & 92.9 \\
		$x_j-x_i,x_j,\lambda_j-\lambda_i,\lambda_j,v_j-v_i,v_j$& 30   & 92.4 \\
		
		\hline
	\end{tabular}
	\caption{The result (\%) of six intuitive input features. $i$ denotes index of anchor point and $j$ denotes its neighbors' indices ($x$: coordinates,$x$:shape context, $\lambda$: eigenvalues, $v$: eigenvectors, $d$: Euclidean distance).}
	\label{input_feature}
	
\end{table}

\begin{figure}[t]
	\centering
	\includegraphics[width=1\columnwidth]{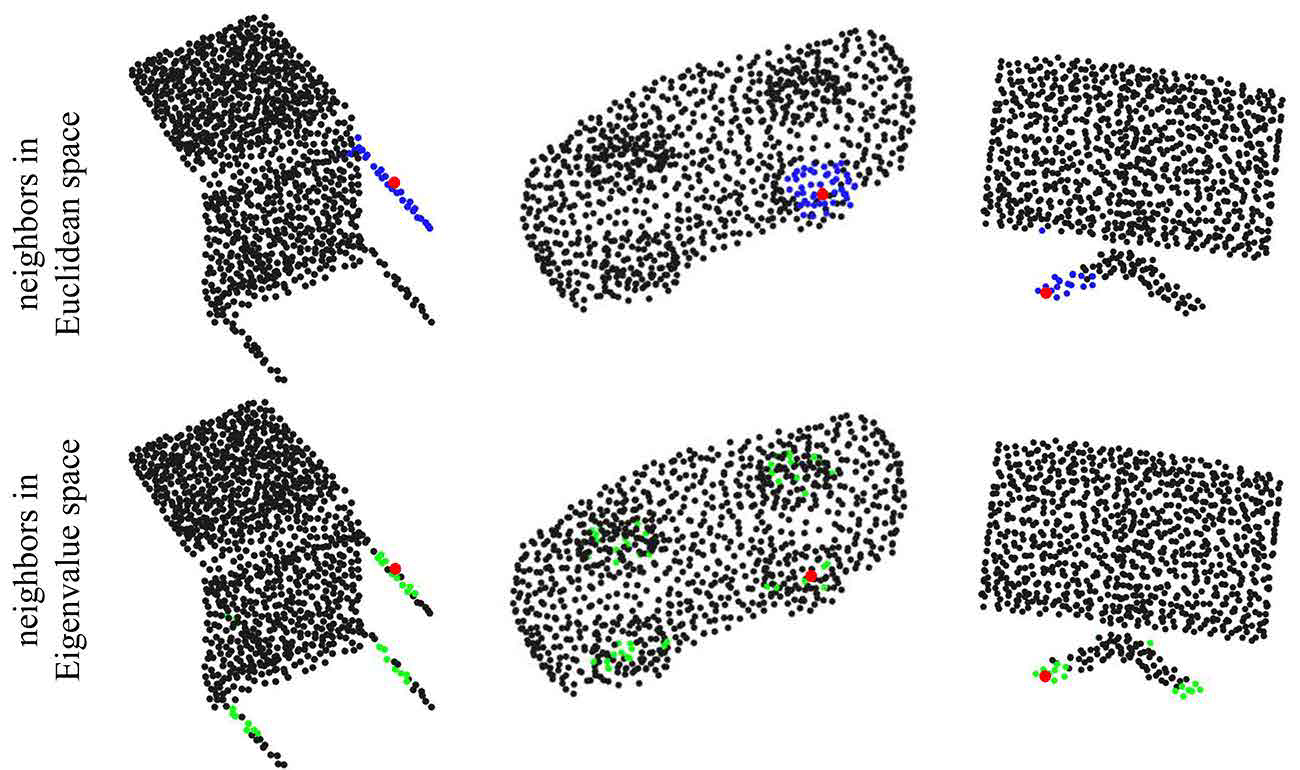} 
	\caption{Visualization of the anchor points(red)'  neighbors in Euclidean space and Eigenvalue space. The neighbors in Euclidean space are colored blue and the neighbors in Eigenvalue space are colored green. The green points have similar local geometry with the corresponding red point.}
	\label{vis_k2}
\end{figure}



\textbf{Input Features.} 
The input features directly affect the representation of local geometry and relations between points, thus how to define the input features is an worth exploring issue. In order to find the most suitable feature combination, we experiment with six settings, whose results are summarized in Table  \ref{input_feature}. As can be seen, using only coordinates, the accuracy can also reach 92.5\%; Inspire by \cite{Xie_2018_CVPR} we use only shape context as the input feature of the points and the result can reach 91.9\%; using the differences of coordinates, the result can reach 92.6\%; with the combination of coordinates and their differences, the result improves to 92.7\%; then we add eigenvalues of points and their differences to the input features, it gets an accuracy of 92.8\%; on this basis, we add 3D Euclidean distance of points and their neighbors, it obtains the accuracy of 92.9\%; however, with the addition of eigenvectors, it can not perform as well as other settings.

\begin{table}
	\centering
	\scriptsize
	\begin{tabular}{p{80pt}lll}
		\hline
		Method & Model  & Forward  & Accuracy\\
		& Size(MB) & Time(ms) & (\%) \\
		\hline
		PointNet\cite{qi2017pointnet}   & 13.4    & 30 & 89.2\\
		PointNet++\cite{qi2017pointnet++} & 7.0     & 603 & 91.9\\
		DGCNN\cite{wang2018dynamic}      & 7.2     & 73  & 92.2\\
		Ours       & 6.0     & 126 & 92.9\\
		\hline
	\end{tabular}
	\caption{Complexity analysis of GS-Net in classification.}
	\label{Complexity_Analysis}
	
\end{table}

\subsection{Complexity Analysis
\label{Sec_ComplexityAnalysis}}
We evaluate the model complexity in terms of model size and forward time in Table \ref{Complexity_Analysis}. The forward time is recorded with a batch size of 8 on a single GTX 1080 GPU, which is the same hardware environment of the comparison models. These models are implemented by Pytorch. As illustrated, our method has the competitive performance with great parameter-efficiency and acceptable speed. 

\subsection{Visualization of GS-Net
	\label{vis_GSNet}}
As shown in Figure \ref{vis_k2}, we visualize the Eigen-Graph of the anchor points (red) from three point clouds.
The blue points in the first row represent the nearest neighbors in Euclidean space, while the green points in the second row indicate the nearest neighbors in Eigenvalue space. As can be seen, the green points have similar local geometry with the anchor point. Moreover, the Eigen-Graph is rotation invariant, as Figure \ref{rotate_k2} shows, nearest neighbors in Eigenvalue space can not be influenced by rotations and translations of the point cloud. 

\section{Conclusion}
We develop Geometry Sharing Net (GS-Net) for point cloud analysis. The core to GS-Net is GSC module, which can share the similar geometric information with distant points and can be integrated into different existing pipelines for point cloud analysis. Moreover, the Eigen-Graph of GSC module improves the rotation and translation robustness fundamentally. 
Experiments have shown that GS-Net achieves the state-of-the-art performance and has robustness to geometric transformations. 

\section{Acknowledgments}
This work is partially supported by the National Key Research and Development Program of China (No. 2016YFC1400704), and National Natural Science Foundation of China (61876176, U1713208), Shenzhen Basic Research Program (JCYJ20170818164704758, CXB201104220032A), the Joint Lab of CAS-HK, Shenzhen Institute of Artificial Intelligence and Robotics for Society.

{\small
	\bibliographystyle{aaai}
	\bibliography{bibFile}

\begin{thebibliography}{}

\bibitem[\protect\citeauthoryear{Atzmon, Maron, and
  Lipman}{2018}]{Atzmon2018Point}
Atzmon, M.; Maron, H.; and Lipman, Y.
\newblock 2018.
\newblock Point convolutional neural networks by extension operators.
\newblock {\em ACM Transactions on Graphics} 37(4):1--12.

\bibitem[\protect\citeauthoryear{Chu, Samari, and Siddiqi}{2018}]{Chu2018Local}
Chu, W.; Samari, B.; and Siddiqi, K.
\newblock 2018.
\newblock Local spectral graph convolution for point set feature learning.

\bibitem[\protect\citeauthoryear{Demantke \bgroup et al\mbox.\egroup
  }{2011}]{demantke2011dimensionality}
Demantke, J.; Mallet, C.; David, N.; and Vallet, B.
\newblock 2011.
\newblock Dimensionality based scale selection in 3d lidar point clouds.
\newblock {\em Int. Arch. Photogramm. Remote Sens. Spat. Inf. Sci} 38(5):W12.

\bibitem[\protect\citeauthoryear{Feng \bgroup et al\mbox.\egroup
  }{2018}]{feng2018gvcnn}
Feng, Y.; Zhang, Z.; Zhao, X.; Ji, R.; and Gao, Y.
\newblock 2018.
\newblock Gvcnn: Group-view convolutional neural networks for 3d shape
  recognition.
\newblock In {\em Proceedings of the IEEE Conference on Computer Vision and
  Pattern Recognition},  264--272.

\bibitem[\protect\citeauthoryear{Gadelha, Rui, and
  Maji}{2018}]{Gadelha2018Multiresolution}
Gadelha, M.; Rui, W.; and Maji, S.
\newblock 2018.
\newblock Multiresolution tree networks for 3d point cloud processing.

\bibitem[\protect\citeauthoryear{Groh, Wieschollek, and
  Lensch}{2018}]{Groh2018Flex}
Groh, F.; Wieschollek, P.; and Lensch, H. P.~A.
\newblock 2018.
\newblock Flex-convolution (million-scale point-cloud learning beyond
  grid-worlds).

\bibitem[\protect\citeauthoryear{Guerrero \bgroup et al\mbox.\egroup
  }{2018}]{guerrero2018pcpnet}
Guerrero, P.; Kleiman, Y.; Ovsjanikov, M.; and Mitra, N.~J.
\newblock 2018.
\newblock Pcpnet learning local shape properties from raw point clouds.
\newblock In {\em Computer Graphics Forum}, volume~37,  75--85.
\newblock Wiley Online Library.

\bibitem[\protect\citeauthoryear{Hornik}{1991}]{hornik1991approximation}
Hornik, K.
\newblock 1991.
\newblock Approximation capabilities of multilayer feedforward networks.
\newblock {\em Neural networks} 4(2):251--257.

\bibitem[\protect\citeauthoryear{Hua, Tran, and Yeung}{2017}]{Hua2017Point}
Hua, B.~S.; Tran, M.~K.; and Yeung, S.~K.
\newblock 2017.
\newblock Point-wise convolutional neural network.

\bibitem[\protect\citeauthoryear{Huang, Wang, and
  Neumann}{2018}]{huang2018recurrent}
Huang, Q.; Wang, W.; and Neumann, U.
\newblock 2018.
\newblock Recurrent slice networks for 3d segmentation of point clouds.
\newblock In {\em Proceedings of the IEEE Conference on Computer Vision and
  Pattern Recognition},  2626--2635.

\bibitem[\protect\citeauthoryear{Klokov and Lempitsky}{2017}]{klokov2017escape}
Klokov, R., and Lempitsky, V.
\newblock 2017.
\newblock Escape from cells: Deep kd-networks for the recognition of 3d point
  cloud models.
\newblock In {\em Proceedings of the IEEE International Conference on Computer
  Vision},  863--872.

\bibitem[\protect\citeauthoryear{Landrieu and
  Simonovsky}{2018}]{landrieu2018large}
Landrieu, L., and Simonovsky, M.
\newblock 2018.
\newblock Large-scale point cloud semantic segmentation with superpoint graphs.
\newblock In {\em Proceedings of the IEEE Conference on Computer Vision and
  Pattern Recognition},  4558--4567.

\bibitem[\protect\citeauthoryear{Li, Chen, and Hee~Lee}{2018}]{li2018so}
Li, J.; Chen, B.~M.; and Hee~Lee, G.
\newblock 2018.
\newblock So-net: Self-organizing network for point cloud analysis.
\newblock In {\em Proceedings of the IEEE conference on computer vision and
  pattern recognition},  9397--9406.

\bibitem[\protect\citeauthoryear{Maturana and
  Scherer}{2015}]{Maturana2015VoxNet}
Maturana, D., and Scherer, S.
\newblock 2015.
\newblock Voxnet: A 3d convolutional neural network for real-time object
  recognition.

\bibitem[\protect\citeauthoryear{Qi \bgroup et al\mbox.\egroup
  }{2017a}]{qi2017pointnet}
Qi, C.~R.; Su, H.; Mo, K.; and Guibas, L.~J.
\newblock 2017a.
\newblock Pointnet: Deep learning on point sets for 3d classification and
  segmentation.
\newblock In {\em Proceedings of the IEEE Conference on Computer Vision and
  Pattern Recognition},  652--660.

\bibitem[\protect\citeauthoryear{Qi \bgroup et al\mbox.\egroup
  }{2017b}]{qi2017pointnet++}
Qi, C.~R.; Yi, L.; Su, H.; and Guibas, L.~J.
\newblock 2017b.
\newblock Pointnet++: Deep hierarchical feature learning on point sets in a
  metric space.
\newblock In {\em Advances in Neural Information Processing Systems},
  5099--5108.

\bibitem[\protect\citeauthoryear{Qi \bgroup et al\mbox.\egroup
  }{2018}]{qi2018frustum}
Qi, C.~R.; Liu, W.; Wu, C.; Su, H.; and Guibas, L.~J.
\newblock 2018.
\newblock Frustum pointnets for 3d object detection from rgb-d data.
\newblock In {\em Proceedings of the IEEE Conference on Computer Vision and
  Pattern Recognition},  918--927.

\bibitem[\protect\citeauthoryear{Rusu \bgroup et al\mbox.\egroup
  }{2008}]{rusu2008towards}
Rusu, R.~B.; Marton, Z.~C.; Blodow, N.; Dolha, M.; and Beetz, M.
\newblock 2008.
\newblock Towards 3d point cloud based object maps for household environments.
\newblock {\em Robotics and Autonomous Systems} 56(11):927--941.

\bibitem[\protect\citeauthoryear{Shen \bgroup et al\mbox.\egroup
  }{2018}]{shen2018mining}
Shen, Y.; Feng, C.; Yang, Y.; and Tian, D.
\newblock 2018.
\newblock Mining point cloud local structures by kernel correlation and graph
  pooling.
\newblock In {\em Proceedings of the IEEE conference on computer vision and
  pattern recognition},  4548--4557.

\bibitem[\protect\citeauthoryear{Simonovsky and
  Komodakis}{2017}]{Simonovsky2017Dynamic}
Simonovsky, M., and Komodakis, N.
\newblock 2017.
\newblock Dynamic edge-conditioned filters in convolutional neural networks on
  graphs.

\bibitem[\protect\citeauthoryear{Srivastava \bgroup et al\mbox.\egroup
  }{2014}]{Srivastava2014Dropout}
Srivastava, N.; Hinton, G.; Krizhevsky, A.; Sutskever, I.; and Salakhutdinov,
  R.
\newblock 2014.
\newblock Dropout: a simple way to prevent neural networks from overfitting.
\newblock {\em Journal of Machine Learning Research} 15(1):1929--1958.

\bibitem[\protect\citeauthoryear{Su \bgroup et al\mbox.\egroup
  }{2015}]{su2015multi}
Su, H.; Maji, S.; Kalogerakis, E.; and Learned-Miller, E.
\newblock 2015.
\newblock Multi-view convolutional neural networks for 3d shape recognition.
\newblock In {\em Proceedings of the IEEE international conference on computer
  vision},  945--953.

\bibitem[\protect\citeauthoryear{Thomas \bgroup et al\mbox.\egroup
  }{2018a}]{thomas2018semantic}
Thomas, H.; Goulette, F.; Deschaud, J.-E.; and Marcotegui, B.
\newblock 2018a.
\newblock Semantic classification of 3d point clouds with multiscale spherical
  neighborhoods.
\newblock In {\em 2018 International Conference on 3D Vision (3DV)},  390--398.
\newblock IEEE.

\bibitem[\protect\citeauthoryear{Thomas \bgroup et al\mbox.\egroup
  }{2018b}]{Thomas2018Tensor}
Thomas, N.; Smidt, T.; Kearnes, S.; Yang, L.; Li, L.; Kohlhoff, K.; and Riley,
  P.
\newblock 2018b.
\newblock Tensor field networks: Rotation- and translation-equivariant neural
  networks for 3d point clouds.

\bibitem[\protect\citeauthoryear{Wang \bgroup et al\mbox.\egroup
  }{2018}]{wang2018dynamic}
Wang, Y.; Sun, Y.; Liu, Z.; Sarma, S.~E.; Bronstein, M.~M.; and Solomon, J.~M.
\newblock 2018.
\newblock Dynamic graph cnn for learning on point clouds.
\newblock {\em arXiv preprint arXiv:1801.07829}.

\bibitem[\protect\citeauthoryear{Wu \bgroup et al\mbox.\egroup
  }{2015}]{Wu20153D}
Wu, Z.; Song, S.; Khosla, A.; and Yu, F.
\newblock 2015.
\newblock 3d shapenets: A deep representation for volumetric shapes.
\newblock In {\em IEEE Conference on Computer Vision \& Pattern Recognition}.

\bibitem[\protect\citeauthoryear{Xie \bgroup et al\mbox.\egroup
  }{2018a}]{xie2018attentional}
Xie, S.; Liu, S.; Chen, Z.; and Tu, Z.
\newblock 2018a.
\newblock Attentional shapecontextnet for point cloud recognition.
\newblock In {\em Proceedings of the IEEE Conference on Computer Vision and
  Pattern Recognition},  4606--4615.

\bibitem[\protect\citeauthoryear{Xie \bgroup et al\mbox.\egroup
  }{2018b}]{Xie_2018_CVPR}
Xie, S.; Liu, S.; Chen, Z.; and Tu, Z.
\newblock 2018b.
\newblock Attentional shapecontextnet for point cloud recognition.
\newblock In {\em The IEEE Conference on Computer Vision and Pattern
  Recognition (CVPR)}.

\bibitem[\protect\citeauthoryear{Xu \bgroup et al\mbox.\egroup
  }{2018}]{xu2018spidercnn}
Xu, Y.; Fan, T.; Xu, M.; Zeng, L.; and Qiao, Y.
\newblock 2018.
\newblock Spidercnn: Deep learning on point sets with parameterized
  convolutional filters.
\newblock In {\em Proceedings of the European Conference on Computer Vision
  (ECCV)},  87--102.

\bibitem[\protect\citeauthoryear{Yang \bgroup et al\mbox.\egroup
  }{2019}]{yang2019modeling}
Yang, J.; Zhang, Q.; Ni, B.; Li, L.; Liu, J.; Zhou, M.; and Tian, Q.
\newblock 2019.
\newblock Modeling point clouds with self-attention and gumbel subset sampling.
\newblock In {\em Proceedings of the IEEE Conference on Computer Vision and
  Pattern Recognition},  3323--3332.

\bibitem[\protect\citeauthoryear{Yi \bgroup et al\mbox.\egroup
  }{2016}]{yi2016scalable}
Yi, L.; Kim, V.~G.; Ceylan, D.; Shen, I.; Yan, M.; Su, H.; Lu, C.; Huang, Q.;
  Sheffer, A.; Guibas, L.; et~al.
\newblock 2016.
\newblock A scalable active framework for region annotation in 3d shape
  collections.
\newblock {\em ACM Transactions on Graphics (TOG)} 35(6):210.

\bibitem[\protect\citeauthoryear{Zaheer \bgroup et al\mbox.\egroup
  }{2017}]{Zaheer2017Deep}
Zaheer, M.; Kottur, S.; Ravanbakhsh, S.; Poczos, B.; Salakhutdinov, R.; and
  Smola, A.
\newblock 2017.
\newblock Deep sets.

\end{thebibliography}
}

\end{document}